\def\year{2020}\relax
\documentclass[letterpaper]{article} 
\usepackage{aaai20}  
\usepackage{times}
\usepackage{latexsym}
\usepackage{graphicx}
\usepackage{amsmath}
\usepackage{makecell}
\usepackage{url}
\usepackage{color}
\usepackage{booktabs}
\usepackage{colortbl}
\usepackage{amssymb}
\usepackage{xspace}
\usepackage{soul}
\usepackage{CJKutf8}
\usepackage{subfigure}
\usepackage{mathtools}
\usepackage{xcolor}
\usepackage{algorithm}
\usepackage{algorithmic}

\newcommand{\dubbelop}{$^{\blacktriangle}$}

\newcommand{\dubbelneer}{$^{\blacktriangledown}$}

\usepackage{xspace}
\usepackage{xcolor}
\usepackage{colortbl}
\usepackage{mathtools}
\usepackage{booktabs}
\newcommand{\cbkgrnd}{\cellcolor{blue!15}}
\usepackage{amsmath}
\usepackage{graphicx} 
\usepackage{graphicx}  
\title{RPM-Oriented Query Rewriting Framework for \\ E-commerce Keyword-Based Sponsored Search \small{(Student Abstract)} }
\author{
Xiuying Chen$^{1,2}$, 
Daorui Xiao$^{3}$, 
Shen Gao$^{2}$, 
Guojun Liu$^{3}$, 
Wei Lin$^{3}$, 
Bo Zheng$^{3}$, 
Dongyan Zhao$^{1,2}$,  
Rui Yan$^{1,2}$\thanks{\;\;Corresponding author: Rui Yan (ruiyan@pku.edu.cn)} \\
$^{1}$Center for Data Science, Peking University, Beijing, China, \\
$^{2}$Wangxuan Institute of Computer Technology, Peking University, China \quad
$^{3}$Alibaba Group, China\\ 
Email: xy-chen@pku.edu.cn}
 
\begin{document}

\maketitle

\begin{abstract}
Sponsored search optimizes revenue and relevance, which is estimated by Revenue Per Mille (RPM).
Existing sponsored search models are all based on traditional statistical models, which have poor RPM performance when queries follow a heavy-tailed distribution.
Here, we propose an RPM-oriented Query Rewriting Framework (RQRF) which outputs related bid keywords that can yield high RPM.
RQRF embeds both queries and bid keywords to vectors in the same implicit space, converting the rewriting probability between each query and keyword to the distance between the two vectors.
For label construction, we propose an RPM-oriented sample construction method, labeling keywords based on whether or not they can lead to high RPM.
Extensive experiments are conducted to evaluate performance of RQRF.
In a one month large-scale real-world traffic of e-commerce sponsored search system, the proposed model significantly outperforms traditional baseline.
\end{abstract}

\section{Introduction}
\label{sec:intro}
Web search seeks to optimize relevance, while sponsored search optimizes RPM, which takes both relevance and bid price together into consideration.
Obviously, it is not possible for advertisers to identify and collect all relevant bid keywords to target their ads.
A common mechanism to solve this problem is query rewriting, which outputs a list of bid keywords (rewrites) that are related to a given query.
Query rewriting has been extensively studied in the context of traditional web search.
Initial studies only focuses on the relevance between a query and bid keyword, without considering the traffic allocation mechanism on the bid keyword side to maximize revenue.
More comprehensive works are based on graphs.
For example, \cite{malekian2008optimizing} formulates a family of graph covering problems whose goals are to suggest a subset of ads with the maximum benefit by suggesting rewrites for a given query.
However, these models also have shortcomings: 
(1) Each query is regarded as a non-splittable graph node, which hinders the full understanding of queries. 
(2) They are all essentially statistic-based memory models, and thus cannot handle heavy-tailed ads.

To overcome obstacles mentioned above, we come up with an RPM-oriented Query Rewrite Framework (RQRF).
RQRF embeds both queries and bid keywords to vectors in the same implicit space.
Specifically, we combine a modified version of a convolutional neural network (CNN) with a self-attention mechanism to learn representations of queries and bid keywords.
For the label construction, since users click on ads instead of the bid keywords, we propose an RPM-oriented sampling method to bridge the gap between labeled ads and unlabeled keywords.
\section{Model}
\label{sec:formulation}
\begin{figure}[htbp]
    \centering
    \begin{minipage}[t]{0.2\textwidth}
        \centering
        \includegraphics[width=3.5cm]{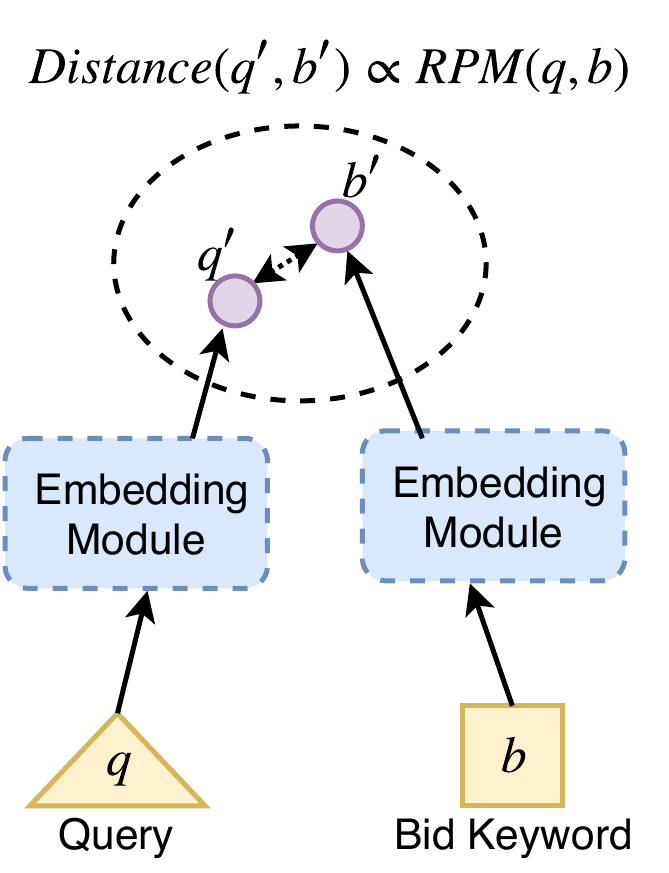}
        \caption{Overview \\ of RQRF }
        \label{schematic}
    \end{minipage}
    \begin{minipage}[t]{0.2\textwidth}
        \centering
        \includegraphics[width=4.2cm]{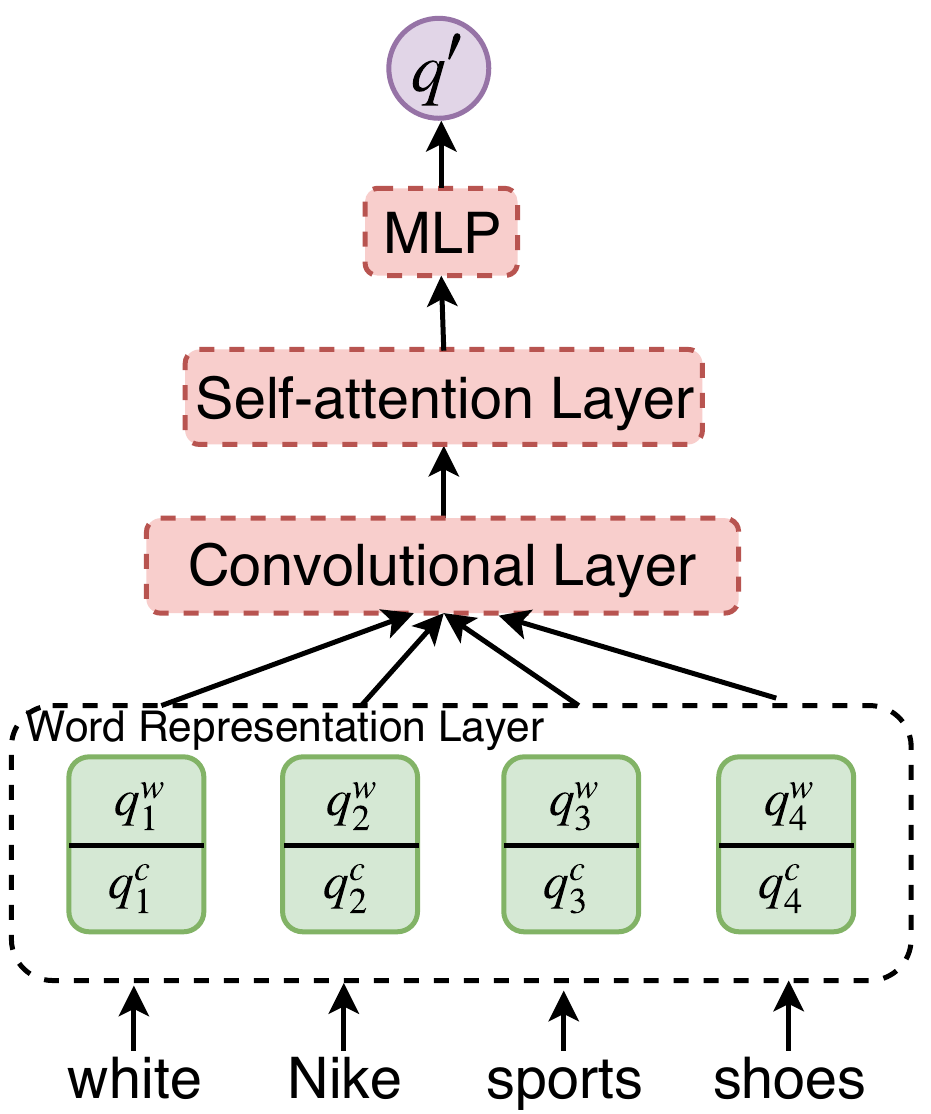}
        \caption{Illustration of the Embedding Module.}
        \label{illustration}
    \end{minipage}
\end{figure}


\noindent \textbf{Embedding Module.}
As shown in Figure 1, our model maps each query-bidword pair to a low dimensional vector $q'$ and $b'$ respectively.
We first introduce the embedding module using query as an example, as shown in Figure~\ref{illustration}.
Bid keywords are embedded in a similar way.
We adopt standard techniques to obtain the embedding of each word $q_{t}$ by concatenating its word embedding $q^{w}_{t}$ and character embedding $q^{c}_{t}$.
The output of a given word $q_{t}$ from this layer is calculated as $m_{t}=[q^{w}_{t}, q^{c}_{t}]$, where ``[,]'' denotes the concatenation between vectors.

Next, a novel deep convolutional neural network architecture from \cite{chollet2017xception} and the self-attention mechanism are adopted, producing $\alpha_t$.
In this way, the model learns the interactions and exchanges of information between words by itself.
Finally, a fully-connected layer is applied to obtain the final query embedding:
\begin{eqnarray}
q'=FC([a_{1},a_{2},...,a_{T}]).
\end{eqnarray}

\noindent\textbf{Label Construction.}
Since there are no click labels on bid keywords, only click labels on ads.
Herein, we propose an RPM-oriented sampling method to construct limited number of training samples, which greatly reduces the size of training data and mitigates the Matthew effect. 
Let the number of ads that bid on keyword $b_{i}$ be $n_{b_{i}}$, and the price ad $a$ bid on $b_i$ be $\text{price}(b_{i}|a)$ for a sample $<q,a>$.
The probability of keyword $b_i$ being positively sampled, i.e. $\text{score}(b_i|q,a)$ is calculated as:
\begin{gather*}
\begin{aligned}
&\operatorname { score } ( b_{i} | q , a ) = \frac { \operatorname { price } ( b_{i} | a )\times \text{relevance}(b_i,q) } { \log \left( n _ { b_{i} } + 1 \right) },
\end{aligned}
\end{gather*}
where $\text{relevance}(b_i,q)=\text{cosine}(b_i,q)$.
Note that the relevance is calculated by pre-trained word embeddings instead of embedding in our model.
We add factor $n_{b_{i}}$ to allow the model to choose keywords that are not that popular, in order to balance advertisement competition.
As for the $\text{relevance}(b_i,q)$ factor, it is used to enhance the importance of relevance between query and bid keyword.
Then the scores of all keywords bought by advertisement $a$ are normalized and we obtain the final sample probability $p ( b_{i} | a , q ) $. 
where $B(a)$ represents all keywords bought by $a$. 
This sampling method leads to a positive correlation between the probability that $q$ matches $b_{i}$ and the RPM value. 
The details of derivation are attached in Appendix.

After obtaining the value of $p$, we use $p$ as a sampling probability to find positive keyword samples, and $1-p$ as a sampling probability to obtain negative samples.
The training objective is to maximize the cosine distance between a query embedding and positive keyword embeddings.


\section{Experiments}
\textbf{Dataset}. We use the search logs from one of the largest e-commerce websites in China to train RQRF.
Overall, there are 9.8 billion training cases and 66 million evaluation and 66 million  test cases.
On average, there are 3.94 words and 8.66 characters in queries and 2.88 words and 6.34 characters in bid keywords.

\noindent \textbf{Offline Evaluation}
Apart from NLL, we also employ commonly-used ranking metrics, including mean average precision (MAP), mean reciprocal rank (MRR) and normalized discounted cumulative gain (NDCG) to evaluate RQRF.
We compare our query embedding model with various baselines, including \textbf{BM25}, \textbf{PM}(non-linear projection),\textbf{CLSM}\cite{DBLP:journals/corr/abs-1812-00158},\textbf{BIDAF}\cite{seo2016bidirectional}.
The main results of our experiments are summarized in Table ~\ref{tab:baseline} where RQRF outperforms all baselines. 
Statistical significance of observed differences between the performance of two runs are tested using a two-tailed paired t-test and is denoted using \dubbelop (or \dubbelneer) for strong significance for $\alpha = 0.01$.


\begin{table}[t]
\small
    \centering
    \caption{Comparison with query embedding baselines. 
        \label{tab:baseline}}
    \begin{tabular}{@{}lcccc@{}}
        \toprule
        Models& NLL & MAP& MRR & NDCG \\
        \midrule
        BM25 & 0.870 & 0.149 & 0.009 &0.157\\
        PM & 0.662 &0.166 & 0.011 & 0.191\\
        BIDAF & 0.623 & 0.180& 0.012 & 0.201\\
        \cbkgrnd CLSM & \cbkgrnd 0.631 & \cbkgrnd 0.191 & \cbkgrnd 0.014 & \cbkgrnd 0.226\\
        RQRF & \textbf{0.617\dubbelneer }& \textbf{0.207 \dubbelop}& \textbf{0.015\dubbelop} & \textbf{0.253\dubbelop}\\
        \bottomrule
    \end{tabular}
\end{table}


\textbf{Online Evaluation.}
RQRF is deployed on a real-world commercial search engine to serve ads as part of A/B testing. 
The traditional statistic-based memory model from \cite{malekian2008optimizing} is used as the control model and RQRF as the treatment.
The flight tests are conducted for a period of one month.
The online results are summarized in Table~\ref{tab:oneline}.
Our model outperforms the baseline in the head traffic by 8.79\% in RPM value.
For tail queries, our model achieves consistently better performance, outperforming the baseline by 16.44\%. 
This demonstrates that our model is especially effective for tail queries.
The ads retrieved by RQRM account for 24.7\% of the total queries in the online test, and bring about 3.07\% RPM increase in total traffic.

\begin{table}[ht]
\small
    \label{tab:oneline}
    \caption{RPM lift rate result on head and tail of query distribution. }
    \label{tab:oneline}
    \centering
    \begin{tabular}{lc}
        \toprule
        Traffic type&  RPM Lift Rate\\
        \midrule
        Head &8.79\%  \\
        Tail  &  16.44\% \\
        Head\&Tail  & 12.75\% \\
        \bottomrule
    \end{tabular}
\end{table}

\section{Conclusion}
In this paper, we propose an RQRM query rewriting model that outputs related bid keywords for a given query.
We are the first to apply a deep neural network model in a keyword-based sponsored search system.
Extensive experiments are conducted to evaluate the performance of RQRF.

\section{Acknowledgments}
This work was supported by the National Key R\&D Program of China (No. 2017YFC0804001), the National Science Foundation of China (NSFC No. 61876196 and NSFC No. 61672058). Rui Yan was sponsored by Alibaba Innovative Research (AIR) Grant.

\bibliography{reference}

\begin{thebibliography}{}

\bibitem[\protect\citeauthoryear{Ahsan and
  Agrawal}{2018}]{DBLP:journals/corr/abs-1812-00158}
Ahsan, H., and Agrawal, R.
\newblock 2018.
\newblock Approximating categorical similarity in sponsored search relevance.
\newblock {\em arXiv preprint arXiv:1812.00158}.

\bibitem[\protect\citeauthoryear{Chollet}{2017}]{chollet2017xception}
Chollet, F.
\newblock 2017.
\newblock Xception: Deep learning with depthwise separable convolutions.
\newblock {\em arXiv preprint}  1610--02357.

\bibitem[\protect\citeauthoryear{Malekian \bgroup et al\mbox.\egroup
  }{2008}]{malekian2008optimizing}
Malekian, A.; Chang, C.-C.; Kumar, R.; and Wang, G.
\newblock 2008.
\newblock Optimizing query rewrites for keyword-based advertising.
\newblock In {\em ACM EC},  10--19.
\newblock ACM.

\bibitem[\protect\citeauthoryear{Seo \bgroup et al\mbox.\egroup
  }{2016}]{seo2016bidirectional}
Seo, M.; Kembhavi, A.; Farhadi, A.; and Hajishirzi, H.
\newblock 2016.
\newblock Bidirectional attention flow for machine comprehension.
\newblock {\em arXiv preprint arXiv:1611.01603}.

\end{thebibliography}
\bibliographystyle{aaai}
\newpage
\section{Appendix}
\label{sec:Appendix}

\section{Dataset}
We use the search logs from one of the largest e-commerce websites in China to train RQRF.
For each query, we use the RPM-sampling method to obtain corresponding positive and negative keywords.
The positive keyword number is the click number of each query, and negative keyword number is four times the positive number.
Since matched ads will be clicked more times, thus will generate more query-ads pairs.
In this way, RQRF implicitly learns to relate queries to keywords that bought by matched ads.
It is worth noting that in practice, each query belongs to one category, and only keywords belonging to this category will be retrieved as candidates to be sampled.
Evaluations are also conducted among bid keywords in the corresponding category.
Overall, there are 9.8 billion training cases and 66 million evaluation and test cases.
On average, there are 3.94 words and 8.66 characters in queries and 2.88 words and 6.34 characters in bid keywords.
We employ the distributed Tensorflow machine learning platform\footnote{\url{https://www.tensorflow.org/}} deployed on a large-scale computing cluster to train our model.

\section{Ablation Study}
In order to explore the impact of the internal structure of RQRF, we conduct an ablation study in Table~\ref{tab:ablation}. 
RQRF-CNN, RQRF-Attention, RQRF-MLP represent RQRF without CNN, attention, and MLP module, respectively.
The performances of all ablation models are worse than that of RQRF, which demonstrates the preeminence of RQRF. 
Specifically, the module that makes the greatest contribution to RQRF is the convolutional layer, without which MAP, MRR, and NDCG drops by 15.9\%, 10\%, and 23.3\%, respectively.

\begin{table}[h]
\small
    \centering
    \caption{Performance of different ablation models.
        \label{tab:ablation}}
    \begin{tabular}{@{}lcccc@{}}
        \toprule
        Models & NLL & MAP & MRR & NDCG\\
        \midrule
        RQRF-CNN& 0.653& 0.174 & 0.014& 0.194\\
        RQRF-Attention &0.676 &  0.177& 0.011 &0.197  \\
        RQRF-MLP &0.662 & 0.176 & 0.012 &0.194\\
        RQRF & \textbf{0.617}& \textbf{0.207 }& \textbf{0.015} & \textbf{0.253}\\
        \bottomrule
    \end{tabular}
    \vspace{-1mm}
\end{table}

\section{Deduction}
\label{sec:Appendix}

Remember that the sample score is calculated as:
\begin{gather*}
\begin{aligned}
\operatorname { score } ( b_{i} | q , a ) &=  \operatorname { price } ( b_{i} | a )\times \frac { \text{relevance}(b_i,q) } { \log \left( n _ { b_{i} } + 1 \right) } \\
&\propto \text{price}(b_i|a) \times f(b_i,q),\\
\text{where } f(b_i,q) &=\frac{\text{relevance}(b_i,q)}{\text{log}(n_{b_i}+1)},\\
\text{relevance}(b_i,q)&=\text{cosine}(b_i,q)
\end{aligned}
\end{gather*}
Here we provide a brief derivation to show that this sampling method leads to a positive correlation between the probability that $q$ matches $b_{i}$ and the RPM value.
RPM estimates earnings that accrue for every 1000 impressions received and is calculated as:
\begin{align}
\text{RPM}(q,b_i)&=\frac{\text{revenue}(q,b_i)}{\text{request}(q)}\\
&= \frac{\textstyle \sum^{n_{b_{i}}}_{j=1}{\left(\text{price} \left (a_{j},b_{i}\right)\times \text{click} \left(a_{j},q\right)\right)}}{\text{request}\left (q\right )}\\  
&=\textstyle \sum^{n_{b_{i}}}_{j=1} \left (\text{price} \left(a_{j},b_{i} \right) \times \frac{\text{click}\left(a_{j},q\right)}{\text{request}\left(q\right)}\right),
\end{align}
where request($q$) is the request amount of query $q$, click($a_j,q$) is the click number of ad $a_j$ for query $q$.
The rewriting probability is calculated as:
\begin{align}
&p\left(b_{i}|q\right)=\textstyle \sum^{n_{b_{i}}}_{j=1}{p\left(b_{i}|a_{j},q\right) \times p\left(a_{j}|q\right)}\\
&= \textstyle \sum^{n_{b_{i}}}_{j=1}{f\left(b_i,q\right) \times \text{price}\left(a_{j},b_{i}\right) \times \frac{\text{click}\left(a_j,q\right)}{\text{request}\left(q\right)}} \\
&=f\left(b_i,q\right) \times \textstyle \sum^{n_{b_{i}}}_{j=1} \left(\text{price} \left(a_j,b_{i}\right) \times \frac{\text{click}\left(a_{j},q\right)}{\text{request}\left(q\right)}\right)\\
&= f\left(b_i,q\right) \times \operatorname {RPM}\left(q,b_i\right)
\end{align}
Above derivation leads to :
\begin{equation}
p ( b_{i} | q ) \propto \operatorname { RPM } ( q,b_{i}  )
\end{equation}
That is, the probability that $q$ matches $b_{i}$ is proportional to the expectation of RPM, and inversely proportional to the number of advertisements bid on $b_{i}$.

\end{document}